\documentclass[final]{cvpr}
\usepackage{anyfontsize}
\usepackage[T1]{fontenc}
\usepackage{lipsum}

\usepackage{times}
\usepackage{epsfig}
\usepackage{graphicx}
\usepackage{amsmath}

\usepackage{amssymb}
\usepackage{dsfont}
\usepackage{bm}
\usepackage{wrapfig}

\DeclareMathOperator*{\argmax}{arg\,max}
\DeclareMathOperator*{\argmin}{arg\,min}

\usepackage{bbding}
\usepackage{booktabs}
\usepackage{multirow}

\usepackage[pagebackref=true,breaklinks=true,colorlinks,bookmarks=false]{hyperref}

\def\cvprPaperID{2728} 
\def\confYear{CVPR 2021}

\begin{document}

\title{Cross-Domain Adaptive Clustering for Semi-Supervised Domain Adaptation}

\author{Jichang Li$^{1}$\quad\quad\quad Guanbin Li$^{2*}$\quad\quad\quad Yemin Shi$^{3}$\quad\quad\quad Yizhou Yu$^{1*}$ \vspace{0mm}\\
$^1$The University of Hong Kong \quad\quad\quad $^2$Sun Yat-sen University \quad\quad\quad $^3$Deepwise AI Lab\\
{\tt\small csjcli@connect.hku.hk, liguanbin@mail.sysu.edu.cn, shiyemin@deepwise.com, yizhouy@acm.org}
 \vspace{0mm}
}
\pagestyle{empty} 

\maketitle
\thispagestyle{empty} 

\newcommand\blfootnote[2]{%
\begingroup
\renewcommand\thefootnote{}\footnote{#1}%
\addtocounter{footnote}{-1}%
\endgroup
}

\blfootnote{*Corresponding Authors.}

\begin{abstract}
In semi-supervised domain adaptation, a few labeled samples per class in the target domain guide features of the remaining target samples to aggregate around them. However, the trained model cannot produce a highly discriminative feature representation for the target domain because the training data is dominated by labeled samples from the source domain. This could lead to disconnection between the labeled and unlabeled target samples as well as misalignment between unlabeled target samples and the source domain. In this paper, we propose a novel approach called Cross-domain Adaptive Clustering to address this problem. To achieve both inter-domain and intra-domain adaptation, we first introduce an adversarial adaptive clustering loss to group features of unlabeled target data into clusters and perform cluster-wise feature alignment across the source and target domains. We further apply pseudo labeling to unlabeled samples in the target domain and retain pseudo-labels with high confidence. Pseudo labeling expands the number of ``labeled" samples in each class in the target domain, and thus produces a more robust and powerful cluster core for each class to facilitate adversarial learning. Extensive experiments on benchmark datasets, including DomainNet, Office-Home and Office, demonstrate that our proposed approach achieves the state-of-the-art performance in semi-supervised domain adaptation.
\end{abstract}
\section{Introduction}

Semi-supervised domain adaptation (SSDA) is a variant of the unsupervised domain adaptation (UDA) problem. With a small number of labeled samples in the target domain, SSDA has the potential to significantly boost performance in comparison to UDA. In general, domain adaptation needs to reduce inter-domain gap (\ie, differences in feature distributions between two domains) and decrease intra-domain gap (\ie, differences among class-wise sub-distributions in the target domain) in order to achieve inter-domain adaptation and intra-domain adaptation simultaneously~\cite{pan2020unsupervised}.

Many existing domain adaptation approaches start with inter-domain adaptation, and guide their models to learn cross-domain sample-wise feature alignment~\cite{shu2018dirt, deng2019cluster, chen2019progressive}, or distribution-wise feature alignment~\cite{gretton2012kernel,long2015learning,kumagai2019unsupervised}. 
In the semi-supervised learning setting, adversarial learning is employed in~\cite{saito2019semi, motiian2017few} to improve sample-wise feature alignment for inter-domain adaptation. However, such previous work ignores extra information indicated by class-wise sub-distributions in the target domain, and thus results in cross-domain feature mismatch during model training, thereby reducing model generalization performance on novel test data in the target domain.

\begin{figure}[t]
\centering
\includegraphics[width=8.0cm,height=3.8cm]{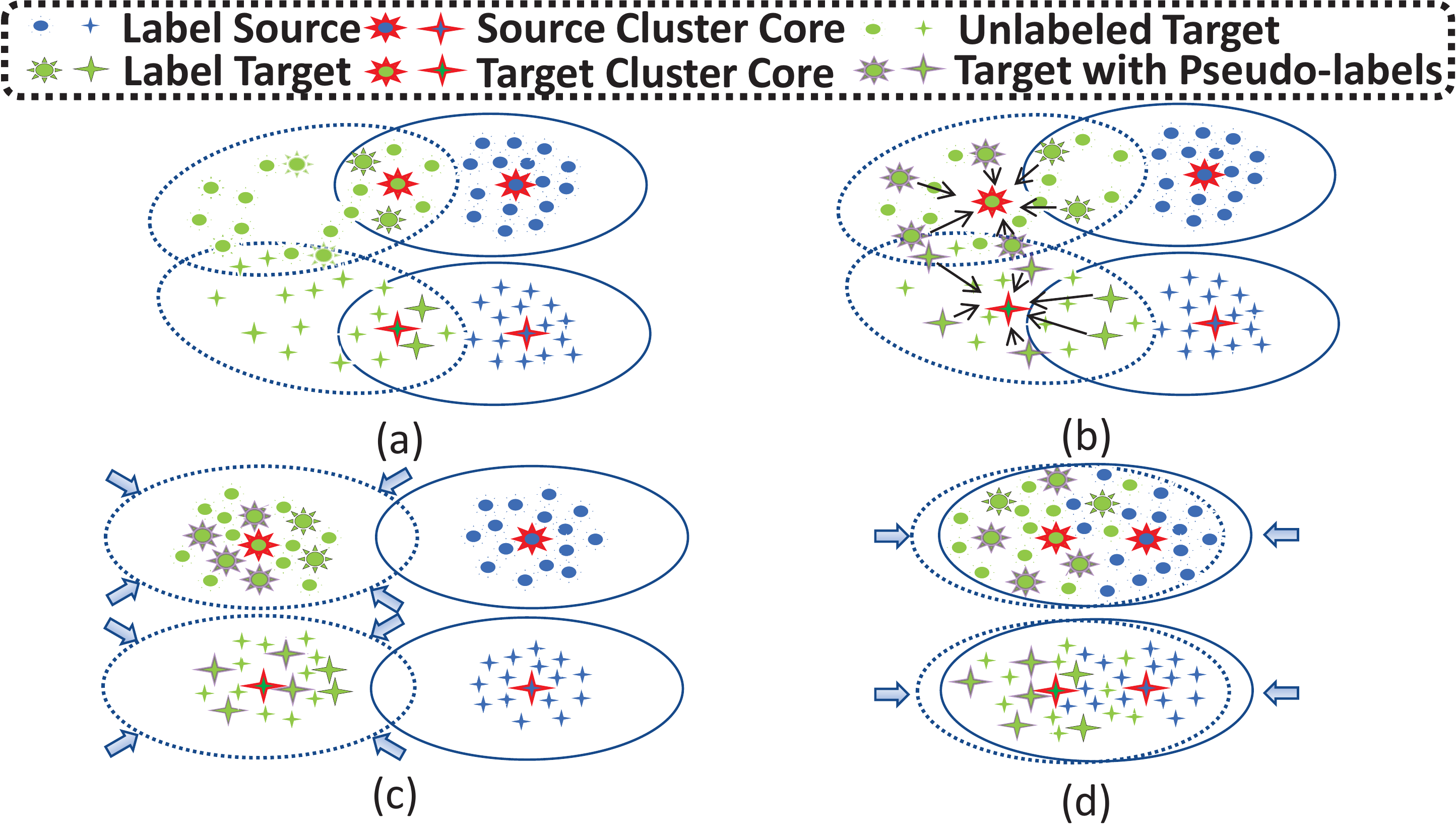} 
\caption{Overview of our Cross-domain Adaptive Clustering (CDAC) approach. (a) Supervision on labeled data from both source and target domains to ensure partial cross-domain feature alignment. (b) Pseudo labeling for giving pseudo-labels on unlabeled target samples to form enhanced target cluster cores with higher robustness and power. (c) Minimization of the adversarial adaptive clustering loss for grouping features from the target domain into clusters. (d) Maximization of the adversarial adaptive clustering loss to facilitate cross-domain cluster-wise feature alignment.}
\label{fig1}
\vspace{-2.0mm}
\end{figure}

Whereafter, much work on domain adaptation has turned to intra-domain adaptation~\cite{kang2019contrastive, gu2020spherical}. By optimizing class-wise sub-distributions within the target domain, the generalization performance of adaptation models can be improved. In the context of semi-supervised domain adaptation, the presence of few labeled target samples is utilized to help features of unlabeled target samples from different classes be guided to aggregate in the corresponding clusters to form perfect class-wise sub-distributions in the target domain, which reduces the possibility of feature mismatch across domains. However, a model trained with supervision on few labeled target samples and labeled source data just can ensure partial cross-domain feature alignment because it only aligns the features of labeled target samples and their correlated nearby ones with the corresponding feature clusters in the source domain~\cite{kim2020attract}. Also, the trained model cannot produce a highly discriminative feature representation for the target domain because the learned feature representation is biased to the sample discrimination of the source domain due to the existence of a much larger scale of labeled samples than those of the target domain~\cite{saito2019semi}. 
These could lead to disconnection between the labeled and unlabeled target samples as well as misalignment between unlabeled target samples and the source domain.

In this paper, we propose a novel approach called {\bf C}ross-{\bf d}omain {\bf A}daptive {\bf C}lustering ({\bf CDAC}), as Figure \ref{fig1} shows, to address the aforementioned problem. It first groups features of unlabeled target data into clusters and further performs cluster-wise feature alignment across the source and target domains rather than sample-wise or distribution-wise feature alignment. In this way, our approach achieves both inter-domain adaptation and intra-domain adaptation simultaneously. More specifically, our proposed approach performs minimax optimization over the parameters of a feature extractor and a classifier. For intra-domain adaptation, the features of unlabeled target samples are guided by labeled target samples to form clusters corresponding to the classes of labeled samples by minimizing an adversarial adaptive clustering loss with respect to the parameters of the feature extractor. For inter-domain adaptation, the classifier is trained to maximize the same loss defined on unlabeled target samples so that cluster-wise feature distribution in the target domain is aligned with the corresponding feature distribution in the source domain.

In addition, we apply pseudo labeling to unlabeled samples in the target domain and retain pseudo-labels with high confidence. In the SSDA setting, since only a very small number (typically one or three) of target samples from each class are labeled, it is hard for such few samples to form a stable and accurate cluster core. Pseudo labeling expands the number of ``labeled" samples in each class in the target domain, and thus produces a more robust and powerful cluster core for each class. Such an enhanced cluster core can attract unlabeled samples from the corresponding class towards itself in the target domain using the adversarial adaptive clustering loss. Therefore, our pseudo labeling technique assists adversarial learning, and helps our SSDA model reach higher performance.

\begin{figure*}[tbp]
\centering
\includegraphics[width=16cm,height=5cm]{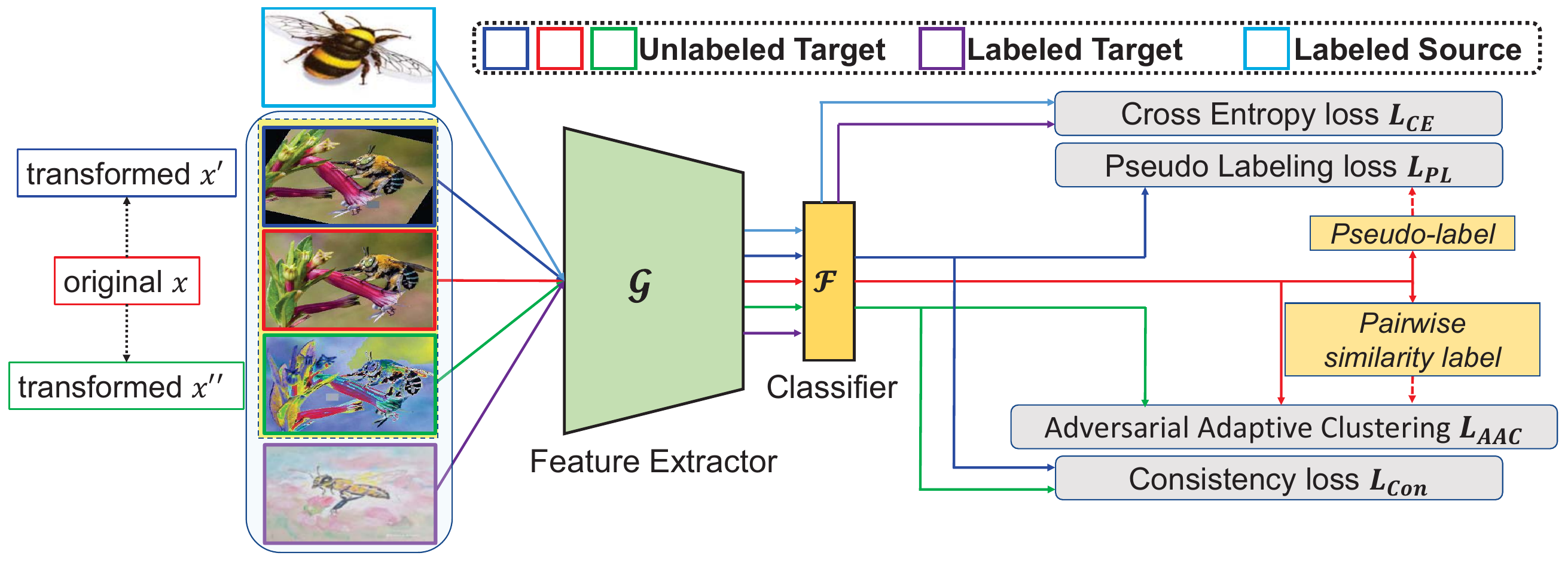}
\caption{Outline of our model architecture and training procedure. Arrows with various colors represent data flows for different types of samples from both source and target domains. The feature extractor $\mathcal{G}$ uses Alexnet or Resnet34 as the backbone network and the classifier $\mathcal{F}$ is an unbiased linear network with a normalized layer, which is shared by both domains. As shown, an image $x$ from unlabeled target data is first fed to the feature extractor and the classifier and then its prediction is constructed to pairwise similarity label and pseudo-label, which are employed as targets for its two different transformed versions, $x'$ and $x''$, to train the model with the adversarial adaptive clustering loss and the proposed pseudo labeling loss, respectively.}
\label{fig2}
\vspace{-2.0mm}
\end{figure*}

In summary, our main contributions of the proposed Cross-domain Adaptive Clustering (CDAC) approach are as follows.
\begin{itemize}
\item	We introduce an adversarial adaptive clustering loss to perform cross-domain cluster-wise feature alignment so as to solve the SSDA problem.
\item	We integrate an adapted version of pseudo labeling to enhance the robustness and power of cluster cores in the target domain to facilitate adversarial learning.
\item	Extensive experiments on benchmark datasets, including \textit{DomainNet}~\cite{peng2019moment}, \textit{Office-Home}~\cite{venkateswara2017deep} and \textit{Office}~\cite{saenko2010adapting}, demonstrate that our proposed CDAC approach achieves the state-of-the-art performance in semi-supervised domain adaptation.
\end{itemize}

\section{Related Work}
\subsection{Adversarial Learning for UDA}
Most domain adaptation algorithms attempt to achieve feature distribution alignment between domains by minimizing the domain shift between the source domain and the target domain, so that the knowledge learned from the source data can be transferred to the target domain and improve its classification performance~\cite{pan2009survey, ghosn2003bias}. Adversarial learning is one of the mainstream solutions~\cite{NEURIPS2018_717d8b3d, Tzeng_2017_CVPR, Cao_2018_ECCV}. Saito {\em et al.}~\cite{saito2018maximum} proposed to train task-specific classifiers and maximize their output discrepancy to detect target samples that are far from the support of the source distribution, then learn to generate target features near the support to fool the classifiers. \cite{vu2019advent, pan2020unsupervised} introduce entropy-based adversarial training to enhance high-confident predictions in the target domain. Moreover, in order to overcome the issue of mode collapse caused by the separate design of task and domain classifiers, Tang {\em et al.}~\cite{tang2020discriminative} proposed discriminative adversarial learning to promote the joint distribution alignment within both feature-level and class-level.

Different from previous sample-wise adversarial learning based domain adaptation methods, we first propose adaptive cluster-wise feature alignment to achieve both inter-domain and intra-domain adaptation. This method can greatly alleviate the situation that the model produces feature representations with bias towards the source domain caused by the dominance of most labeled source samples during model training, and can reduce the difficulty of exploring the decision boundary of the classifier by improving the cohesion of unlabeled samples in the target domain, so as to improve the performance of the model in a two-pronged manner.

\subsection{Pseudo Labeling on UDA}
Pseudo labeling, a.k.a self-training, is often used in semi-supervised learning, aiming to give reliable pseudo-labels to unlabeled data through an ensemble of output predictions from multiple models and assist model training to improve performance and its generalization~\cite{lee2013pseudo, Grandvalet2005Semi}. In the field of semi-supervised image classification, the reliability of pseudo-labels is usually improved by integrating the output predictions of one model with multiple augmented inputs~\cite{berthelot2019mixmatch, sohn2020fixmatch}, outputs of different models~\cite{wu2019mutual}, or multiple predictions of the same model in different training stages~\cite{yang2020pseudo, arazo2020pseudo, feng2020semi}. In previous researches, pseudo labeling is also proved to be effective in domain adaptation, \eg, \cite{pan2020unsupervised} proposed entropy-based ranking function to separate the target domain data into an easy and hard split followed by employing self-supervised adaptation from easy to hard for decreasing intra-domain gap. To avoid introducing noise from pseudo labeling, \cite{gu2020spherical} constructed a robust Gaussian-Uniform mixture model in spherical feature space to guarantee the correctness of given pseudo-labels from unlabeled target data. 

In this work, pseudo labeling is employed to give pseudo-labels for unlabeled target data with high probabilistic confidence and thus expand the number of ``labeled'' samples in each class of the target domain, resulting in a more robust and powerful cluster core for each class to facilitate adversarial learning. 

\subsection{Semi-supervised Domain Adaptation}
Semi-supervised domain adaptation~(SSDA) is a relatively promising form of transfer learning, which intents to leverage a small number of labeled samples~(e.g, one or few samples per class) in the target domain and give full play to their potential to greatly improve the performance of domain adaptation. Recently, SSDA has recently attracted wide attentions~\cite{saito2019semi, qin2020opposite, 2020Bidirectional, li2020online, kim2020attract, yang2020mico} from researchers.~\cite{saito2019semi, qin2020opposite} first proposed to solve SSDA by aligning the features from both domains by means of adversarial learning. \cite{2020Bidirectional} proposed to reduce intra-domain discrepancy within the target domain to attract unaligned target sub-distributions towards the corresponding source sub-distributions so as to improve feature alignment across domains. In addition, \cite{pan2020unsupervised} proposed to decompose SSDA into a semi-supervised learning (SSL) problem in the target domain and an unsupervised domain adaptation (UDA) problem across domains, and then train two classifiers using Mixup and Co-training methods, so as to bridge the gap and exchange expertise between the source and target domains. Furthermore,~\cite{li2020online} proposed to explore the optimal initial weights for the adaptation model using online meta-learning. Most of the previous approaches solve SSDA based on sample-wise feature alignment. In this work, we take an attempt to use adaptive cluster-wise feature alignment affiliated with pseudo labeling to achieve both inter-domain and intra-domain adaptation.

\section{Methodology}
In this section, we first introduce the background and notations of SSDA, and then present our proposed Cross-domain Adaptive Clustering (CDAC) approach, which contains an adversarial adaptive clustering loss and a pseudo labeling loss. Finally, we summarize the overall loss used in our work. An outline of our model architecture and training procedure is shown in Figure \ref{fig2}.

\subsection{Semi-supervised Domain Adaptation}
Semi-supervised domain adaptation seeks a classifier for a target domain when given labeled data $\mathcal{S}=\left(x_i^{s},y_i^{s}\right)_{i=1}^{N_{s}}$ from a source domain as well as both unlabeled data $\mathcal{U}=\left\{\left(x_i^{u}\right)\right\}_{i=1}^{N_{u}}$ and labeled data $\mathcal{L}=\left\{\left(x_i^{l},y_i^{l}\right)\right\}_{i=1}^{N_{l}}$ from the target domain. $\mathcal{S}, \mathcal{U}$ and $\mathcal{L}$ represent three subsets of available data in this problem, and they contain $N_{s}$, $N_{u}$ and $N_{l}$ instances, respectively. In the semi-supervised setting, $N_{l}$ is much smaller than $N_{s}$ and $N_{u}$, and only contains one shot or few shots per class. Each data point $x_i^{s} (x_i^{l})$ from $\mathcal{S} (\mathcal{L})$ has its associated label $y_i^{s} (y_i^{l})$, while any data point $x_i^{u}$ from $\mathcal{U}$ has none. Our work aims to make our SSDA model trained using $\mathcal{S}$, $\mathcal{U}$ and $\mathcal{L}$ perform well on test data from the target domain.

Our network consists of two components, \ie, a feature extractor $\mathcal{G}$, parameterized by $\theta_\mathcal{G}$, and a classifier $\mathcal{F}$, parameterized by $\theta_\mathcal{F}$, as in existing work~\cite{saito2019semi, 2020Bidirectional, kim2020attract}. The classifier $\mathcal{F}$ is an unbiased linear network with a normalization layer, which maps features from the feature extractor $\mathcal{G}$ into a spherical feature space. This similarity-based feature space is more suitable for decreasing the feature variance of samples sharing the same class label~\cite{saito2019semi, kim2020attract}. These are commonly used model settings for the SSDA problem~\cite{saito2019semi, 2020Bidirectional, kim2020attract}.

The feature of an input image $x$, $\mathcal{G}\left(x\right)$, is fed into the classifier $\mathcal{F}$ to obtain the probabilistic prediction as follows:
\begin{equation}
p\left(x\right)=\sigma(\mathcal{F}(\mathcal{G}(x))),
\end{equation}
where $\sigma\left(\cdot\right)$ is the softmax function. For convenience, we often abbreviate $p\left(x\right)$ as $\mathbf{p}$, \ie, $\mathbf{p}=p\left(x\right)$.

To train our model with supervision from all labeled data from both source and target domains, we follow the practices of existing work on SSDA~\cite{saito2019semi, qin2020opposite, 2020Bidirectional, li2020online, kim2020attract, yang2020mico}, and include the following standard cross-entropy loss in the training loss,
\begin{equation}
{\bm{L}}_{\bm{CE}}=-\sum_{\left(x,y\right)\in\mathcal{S}\cup\mathcal{L}}{y \log{\left(p\left(x\right)\right)}}.
\end{equation}

\subsection{Adversarial Adaptive Clustering}
\label{sec_AAC}
The key idea in our work is the introduction of an adversarial adaptive clustering loss into semi-supervised domain adaptation to group features in the target domain into clusters and further perform cross-domain cluster-wise feature alignment to achieve inter-domain adaptation and intra-domain adaptation simultaneously. Underlying assumptions are that features of sample images form clusters and samples from the same cluster should have similar features and share the same class label. This loss first computes pairwise similarities among features of unlabeled samples in the target domain, then forces the class labels predicted by the classifier for such samples with pairwise feature similarities to be consistent. The latter is achieved by training the model with a binary cross-entropy loss, where binary pairwise feature similarities are used as groundtruth labels. This loss can force the features from the target domain to form clusters.

In detail, the above approach requires setting up connections on the basis of a similarity measure between sample pairs ($x_i^{u}$,$x_j^{u}$) from the same mini-batch. According to the above assumption, for a pair of similar samples, we set a pairwise pseudo-label $s_{ij}=1$ (\ie, pairwise connection between paired samples); otherwise, $s_{ij}=0$ for dissimilar samples. According to~\cite{han2020automatically}, pairwise feature similarity can be measured using the indices of feature elements rank ordered according to their magnitudes. If two samples share the same top-$k$ indices in their respective lists of rank ordered feature elements, the paired samples belong to the same class with a high confidence and thus $s_{ij}=1$; otherwise, $s_{ij}=0$. Therefore, we can formulate pairwise similarity label as follows,
\begin{equation}
\label{sij}
s_{ij}= \mathds{1}\{\mbox{top}k\left(\mathcal{G}\left(x_i^{u}\right)\right)=\mbox{top}k\left(\mathcal{G}\left(x_j^{u}\right)\right)\},
\end{equation}
where $\mbox{top}k\left(\cdot\right)$ denotes the top-$k$ indices of rank ordered feature elements and we set $k=5$. And $\mathds{1}\{\cdot\}$ is an indicator function.

Then we establish pairwise comparisons among unlabeled target data using the binary cross-entropy loss, which utilize the above pairwise feature similarity labels of sample pairs in a mini-batch as targets, \ie, our adversarial adaptive clustering loss ${\bm{L}}_{\bm{AAC}}$ can be written as follows, 
\begin{equation}
\label{l_aac}
\begin{split}
{\bm{L}}_{\bm{AAC}}=-\sum_{i=1}^{M}\sum_{j=1}^{M}s_{ij}&\log(\mathbf{\mathrm}\mathbf{p}_i^\mathsf{T}\mathbf{p}_j^\mathbf{\prime})   \\
&+(1-s_{ij})\log(1-\mathbf{\mathrm}\mathbf{p}_i^\mathsf{T}\mathbf{p}_j^\mathbf{\prime}),
\end{split}
\end{equation}
where $M$ is the number of unlabeled target samples in each mini-batch and $\mathbf{p}_i=p\left(x_i^{u}\right)=\sigma(\mathcal{F}(\mathcal{G}(x_i^{u})))$ represents the prediction of an image $x_i^{u}$ in the mini-batch. Also, $\mathbf{p}_i^\mathbf{\prime}=p\left(x_i^\mathbf{\prime}\right)=\sigma(\mathcal{F}(\mathcal{G}(x_i^\mathbf{\prime})))$ indicates the prediction of a transformed image $x_j^\prime$, which is an augmented version of $x_j^{u}$ using a data augmentation technique. The inner product $\mathbf{\mathrm}\mathbf{p}_i^\mathsf{T}\mathbf{p}_j^\mathbf{\prime}$ in Equation (\ref{l_aac}) is used as a similarity score, which predicts whether image $x_i^{u}$ and the transformed version of image $x_j^{u}$ share the same class label or not. Besides, as illustrated in~\cite{rebuffi2020lsd}, data augmentation techniques combined in the process of pairwise comparison can significantly strengthen the model performance.

What is the goal of Cross-domain Adaptive Clustering achieved using the $\bm{L}_{\bm{AAC}}$ loss? 
Similar to~\cite{saito2019semi}, we also enforce supervision on labeled samples from the source and target domains and perform minimax training on unlabeled target domain samples to optimize the model, but we replace the conditional entropy loss with our adversarial adaptive clustering loss. In our work, directly minimizing the $\bm{L}_{\bm{AAC}}$ loss makes features of similar samples in the target domain close but features of dissimilar ones distant so that features form clusters within the target domain. However, the learned feature representation in the target domain would be always biased towards the source domain because a large number of source labels dominate the supervision process. Thus direct minimization of $\bm{L}_{\bm{AAC}}$ over unlabeled target domain data would make this worse and give rise to more severe overfitting. Therefore, we utilize a gradient reversal layer~\cite{ganin2015unsupervised} to flip the gradients of $\bm{L}_{\bm{AAC}}$ between the feature extractor and the classifier and, in this situation, the classifier is still enforced to ensure correct classification in the target domain. In other words, the maximization of $\bm{L}_{\bm{AAC}}$ on unlabeled target domain data would decrease the bias of feature representations towards the source domain and encourage the model to produce more domain-invariant features so as to facilitate cross-domain feature alignment.
Thus, a preliminary loss function for adversarial learning in our network can be summarized as follows,
\begin{equation}
\label{pre_loss}
\begin{aligned}
\theta_{\mathcal{G}}^{\ast} = \mathop{\argmin}_{\theta_\mathcal{G}}{{\bm{L}_{\bm{CE}}}+\lambda{\bm{L}_{\bm{AAC}}}},\\
\theta_{\mathcal{F}}^{\ast} = \mathop{\argmin}_{\theta_\mathcal{F}}{{\bm{L}_{\bm{CE}}}-\lambda{\bm{L}_{\bm{AAC}}}},
\end{aligned}
\end{equation}
where $\lambda$ is a scalar hyper-parameter that controls the balance between the cross-entropy loss and the proposed adversarial adaptive clustering loss.

\subsection{Pseudo Labeling for Unlabeled Target Domain Data}
Due to the small number of labeled target domain samples in the SSDA problem, it is hard for the adversarial adaptive clustering loss to form stable and accurate cluster cores in the target domain during model training, which may negatively affect cross-domain cluster-wise feature alignment. To solve this problem, we apply pseudo labeling to unlabeled target samples and retain pseudo-labels with high confidence to expand the number of ``labeled'' samples in the target domain, thereby forming more robust cluster cores for different classes.
Pseudo labeling is a classic technique for semi-supervised learning~\cite{arazo2020pseudo, sohn2020fixmatch}, and utilizes the prediction capability of a model to generate artificial hard labels for a subset of unlabeled samples and then train the model with a supervised loss involving these artificial labels. In our work, we choose the progressive pseudo labeling technique in~\cite{sohn2020fixmatch}.

In the proposed pseudo labeling process, we first feed an image $x_j^{u}$ from a mini-batch of unlabeled images into the current model, and the prediction $\mathbf{p}_j=p\left(x_j^{u}\right)=\sigma(\mathcal{F}(\mathcal{G}(x_j^{u})))$ from the model is then converted to a one-hot hard label ${\hat{y}}_j^{u}=\mathop{\argmax}{(\mathbf{p}_j)}$, which is used as a pseudo label in a supervised loss. Afterwards, the prediction $\mathbf{p}_j^{\mathbf{\prime}\mathbf{\prime}}=p\left(x_j^{\prime\prime}\right)$ produced from another transformed image $x_j^{\prime\prime}$ for the same image $x_j^{u}$ is obtained to increase the input diversity of our model. 
Therefore, in this section, our model is trained using the standard cross-entropy loss as follows,
\begin{equation}
\label{L_pl}
\bm{L}_{\bm{PL}}=-\sum_{j=1}^{M}{\mathds{1}\{\max({\mathbf{p}}_j)\geq\tau\}}\cdot{\hat{y}}_j^{u}\log(\bm{\bm{p}}\left(x_j^{\prime\prime}\right)),
\end{equation}
where $\mathbf{p}_j^\mathbf{\prime\prime}=p\left(x_j^\mathbf{\prime\prime}\right)=\sigma(\mathcal{F}(\mathcal{G}(x_j^\mathbf{\prime\prime})))$ denotes the model prediction of the transformed image $x_j^{\prime\prime}$, and $\tau$ is a scalar confidence threshold that determines the subset of pseudo labels that should be retained for model training.

Our $\bm{L}_{\bm{PL}}$ loss is employed to enhance the adversarial adaptive clustering loss. Once pseudo-labels with high confidence are identified and used for model training, more robust cluster cores in the target domain can be established to make the feature clusters in the target domain better aligned with the source domain ones.

\subsection{Overall Loss}
The overall loss function for training our SSDA network can be summarized as follows,
\begin{equation}
\label{final_loss}
\begin{aligned}
\theta_{\mathcal{G}}^{\ast} = \mathop{\argmin}_{\theta_{\mathcal{G}}}{{\bm{L}_{\bm{CE}}}+\lambda{\bm{L}_{\bm{AAC}}}+{\bm{L}}_{\bm{PL}}+{\bm{L}}_{\bm{Con}}},\\
\theta_{\mathcal{F}}^{\ast} = \mathop{\argmin}_{\theta_{\mathcal{F}}}{{\bm{L}_{\bm{CE}}}-\lambda{\bm{L}_{\bm{AAC}}}+{\bm{L}}_{\bm{PL}}+{\bm{L}}_{\bm{Con}}},
\end{aligned}
\end{equation}
where
\begin{equation}
\label{L_con}
{\bm{L}}_{\bm{Con}}=w\left(t\right)\sum_{j=1}^{M}{||\mathbf{p}_j^\mathbf{\prime}-\mathbf{p}_j^\mathbf{\prime\prime}||^2},
\end{equation}
and $w\left(t\right)=\nu e^{-5\left(1-\frac{t}{T}\right)^2}$ is a ramp-up function used in~\cite{laine2016temporal} with the scalar coefficient $\nu$, the current time step $t$ and the total number of steps $T$ in the ramp-up process. In order to improve the input diversity of our model, we have created two different transformed versions of each unlabeled image in the target domain to implement the adversarial adaptive clustering loss and the pseudo labeling loss, respectively. Therefore, we employ a consistency loss, ${\bm{L}}_{\bm{Con}}$, to keep the model predictions on these two transformed images consistent.

\begin{table*}[tbp]
\fontsize{9.2pt}{9.2pt} 
\selectfont
\begingroup
\setlength{\tabcolsep}{1.2pt} 
\renewcommand{\arraystretch}{1.08} 
\caption{Accuracy(\%) on \textit{DomainNet} under the settings of 1-shot and 3-shot using Alexnet and Resnet34 as backbone networks.}
\label{base_domainNet_table}
\begin{center}
\begin{tabular}{c|c|cccccccccccccc|cc}
\specialrule{.1em}{.05em}{.05em}
\multirow{2}{*}{Net} & \multirow{2}{*}{Method} & \multicolumn{2}{c}{R$\rightarrow$C} & \multicolumn{2}{c}{R$\rightarrow$P} & \multicolumn{2}{c}{P$\rightarrow$C} & \multicolumn{2}{c}{C$\rightarrow$S} & \multicolumn{2}{c}{S$\rightarrow$P} & \multicolumn{2}{c}{R$\rightarrow$S} & \multicolumn{2}{c|}{P$\rightarrow$R} & \multicolumn{2}{c}{Mean} \\
 & & 1-shot & 3-shot & 1-shot & 3-shot & 1-shot & 3-shot & 1-shot & 3-shot & 1-shot & 3-shot & 1-shot & 3-shot & 1-shot & 3-shot & 1-shot & 3-shot \\ \hline
\multirow{8}{*}{Alexnet} & S+T & 43.3 & 47.1 & 42.4 & 45.0 & 40.1 & 44.9 & 33.6 & 36.4 & 35.7 & 38.4 & 29.1 & 33.3 & 55.8 & 58.7 & 40.0 & 43.4 \\
 & DANN & 43.3 & 46.1 & 41.6 & 43.8 & 39.1 & 41.0 & 35.9 & 36.5 & 36.9 & 38.9 & 32.5 & 33.4 & 53.5 & 57.3 & 40.4 & 42.4 \\
 & ENT & 37.0 & 45.5 & 35.6 & 42.6 & 26.8 & 40.4 & 18.9 & 31.1 & 15.1 & 29.6 & 18.0 & 29.6 & 52.2 & 60.0 & 29.1 & 39.8 \\
 & MME & 48.9 & 55.6 & 48.0 & 49.0 & 46.7 & 51.7 & 36.3 & 39.4 & 39.4 & 43.0 & 33.3 & 37.9 & 56.8 & 60.7 & 44.2 & 48.2 \\
 & Meta-MME & - & 56.4 & - & 50.2 & & 51.9 & - & 39.6 & - & 43.7 & - & 38.7 & - & 60.7 & - & 48.8 \\
 & BiAT & 54.2 & 58.6 & 49.2 & 50.6 & 44.0 & 52.0 & 37.7 & 41.9 & 39.6 & 42.1 & 37.2 & 42.0 & 56.9 & 58.8 & 45.5 & 49.4 \\
 & APE & 47.7 & 54.6 & 49.0 & 50.5 & 46.9 & 52.1 & 38.5 & 42.6 & 38.5 & 42.2 & 33.8 & 38.7 & 57.5 & 61.4 & 44.6 & 48.9 \\
 & CDAC & \textbf{56.9} & \textbf{61.4} & \textbf{55.9} & \textbf{57.5} & \textbf{51.6} & \textbf{58.9} & \textbf{44.8} & \textbf{50.7} & \textbf{48.1} & \textbf{51.7} & \textbf{44.1} & \textbf{46.7} & \textbf{63.8} & \textbf{66.8} & \textbf{52.1} & \textbf{56.2} \\ 
\hline
\hline
\multirow{9}{*}{Resnet34} & S+T & 55.6 & 60.0 & 60.6 & 62.2 & 56.8 & 59.4 & 50.8 & 55.0 & 56.0 & 59.5 & 46.3 & 50.1 & 71.8 & 73.9 & 56.9 & 60.0 \\
 & DANN & 58.2 & 59.8 & 61.4 & 62.8 & 56.3 & 59.6 & 52.8 & 55.4 & 57.4 & 59.9 & 52.2 & 54.9 & 70.3 & 72.2 & 58.4 & 60.7 \\
 & ENT & 65.2 & 71.0 & 65.9 & 69.2 & 65.4 & 71.1 & 54.6 & 60.0 & 59.7 & 62.1 & 52.1 & 61.1 & 75.0 & 78.6 & 62.6 & 67.6 \\
 & MME & 70.0 & 72.2 & 67.7 & 69.7 & 69.0 & 71.7 & 56.3 & 61.8 & 64.8 & 66.8 & 61.0 & 61.9 & 76.1 & 78.5 & 66.4 & 68.9 \\
 & UODA & 72.7 & 75.4 & 70.3 & 71.5 & 69.8 & 73.2 & 60.5 & 64.1 & 66.4 & 69.4 & 62.7 & 64.2 & 77.3 & 80.8 & 68.5 & 71.2 \\
 & Meta-MME & - & 73.5 & - & 70.3 & - & 72.8 & - & 62.8 & - & 68.0 & - & 63.8 & - & 79.2 & - & 70.1 \\
 & BiAT & 73.0 & 74.9 & 68.0 & 68.8 & 71.6 & 74.6 & 57.9 & 61.5 & 63.9 & 67.5 & 58.5 & 62.1 & 77.0 & 78.6 & 67.1 & 69.7 \\
 & APE & 70.4 & 76.6 & 70.8 & 72.1 & 72.9 & 76.7 & 56.7 & 63.1 & 64.5 & 66.1 & 63.0 & 67.8 & 76.6 & 79.4 & 67.6 & 71.7 \\
 & CDAC & \textbf{77.4} & \textbf{79.6} & \textbf{74.2} & \textbf{75.1} & \textbf{75.5} & \textbf{79.3} & \textbf{67.6} & \textbf{69.9} & \textbf{71.0} & \textbf{73.4} & \textbf{69.2} & \textbf{72.5} & \textbf{80.4} & \textbf{81.9} & \textbf{73.6} & \textbf{76.0} \\ 
 \specialrule{.1em}{.05em}{.05em}
 
\end{tabular}
\end{center}
\endgroup
\end{table*}

\begin{table*}[tbp]
\centering
\caption{Accuracy(\%) on \textit{Office-Home} under the setting of 3-shot using Alexnet and Resnet34 as backbone networks.}
\label{base_office_home_table}
\resizebox{\linewidth}{!}{
\begin{tabular}{c|c|cccccccccccc|c}
\specialrule{.1em}{.05em}{.05em}
Net & Method & R$\rightarrow$C & R$\rightarrow$P & R$\rightarrow$A & P$\rightarrow$R & P$\rightarrow$C & P$\rightarrow$A & A$\rightarrow$P & A$\rightarrow$C & A$\rightarrow$R & C$\rightarrow$R & C$\rightarrow$A & C$\rightarrow$P & Mean \\
\hline
\multirow{8}{*}{Alexnet} & S+T & 44.6 & 66.7 & 47.7 & 57.8 & 44.4 & 36.1 & 57.6 & 38.8 & 57.0 & 54.3 & 37.5 & 57.9 & 50.0 \\
 & DANN & 47.2 & 66.7 & 46.6 & 58.1 & 44.4 & 36.1 & 57.2 & 39.8 & 56.6 & 54.3 & 38.6 & 57.9 & 50.3 \\
 & ENT & 44.9 & 70.4 & 47.1 & 60.3 & 41.2 & 34.6 & 60.7 & 37.8 & 60.5 & 58.0 & 31.8 & 63.4 & 50.9 \\
 & MME & 51.2 & 73.0 & 50.3 & 61.6 & 47.2 & 40.7 & 63.9 & 43.8 & 61.4 & 59.9 & 44.7 & 64.7 & 55.2 \\
 & Meta-MME & 50.3 & - & - & - & 48.3 & 40.3 & - & 44.5 & - & - & 44.5 & - & - \\
 & BiAT & - & - & - & - & - & - & - & - & - & - & - & - & 56.4 \\
 & APE & 51.9 & 74.6 & 51.2 & 61.6 & 47.9 & 42.1 & \textbf{65.5} & 44.5 & 60.9 & 58.1 & 44.3 & 64.8 & 55.6 \\
 & CDAC & \textbf{54.9} & \textbf{75.8} & \textbf{51.8} & \textbf{64.3} & \textbf{51.3} & \textbf{43.6} & 65.1 & \textbf{47.5} & \textbf{63.1} & \textbf{63.0} & \textbf{44.9} & \textbf{65.6} & \textbf{56.8} \\
\hline
\hline
\multirow{7}{*}{Resnet34} & S+T & 55.7 & 80.8 & 67.8 & 73.1 & 53.8 & 63.5 & 73.1 & 54.0 & 74.2 & 68.3 & 57.6 & 72.3 & 66.2 \\
 & DANN & 57.3 & 75.5 & 65.2 & 69.2 & 51.8 & 56.6 & 68.3 & 54.7 & 73.8 & 67.1 & 55.1 & 67.5 & 63.5 \\
 & ENT & 62.6 & 85.7 & 70.2 & 79.9 & 60.5 & 63.9 & 79.5 & 61.3 & 79.1 & 76.4 & 64.7 & 79.1 & 71.9 \\
 & MME & 64.6 & 85.5 & 71.3 & 80.1 & 64.6 & 65.5 & 79.0 & 63.6 & 79.7 & 76.6 & 67.2 & 79.3 & 73.1 \\
 & Meta-MME & 65.2 & - & - & - & 64.5 & 66.7 & - & 63.3 & - & - & \textbf{67.5} & - & - \\
 & APE & 66.4 & \textbf{86.2} & \textbf{73.4} & \textbf{82.0} & 65.2 & 66.1 & \textbf{81.1} & 63.9 & 80.2 & 76.8 & 66.6 & 79.9 & 74.0 \\
 & CDAC & \textbf{67.8} & 85.6 & 72.2 & 81.9 & \textbf{67.0} & \textbf{67.5} & 80.3 & \textbf{65.9} & \textbf{80.6} & \textbf{80.2} & 67.4 & \textbf{81.4} & \textbf{74.2} \\
\specialrule{.1em}{.05em}{.05em}
\end{tabular}}
\vspace{-2.0mm}
\end{table*}
\begin{table}[t]
\centering
\caption{Accuracy(\%) on \textit{Office} under the settings of 1-shot and 3-shot on the Alexnet backbone network.}
\label{base_office_table}
\resizebox{\linewidth}{!}{
\begin{tabular}{c|c|cccc|cc}
\specialrule{.1em}{.05em}{.05em}
\multirow{2}{*}{Net} & \multirow{2}{*}{Method} & \multicolumn{2}{c}{W$\rightarrow$A} & \multicolumn{2}{c|}{D$\rightarrow$A} & \multicolumn{2}{c}{Mean} \\
 & & 1-shot & 3-shot & 1-shot & 3-shot & 1-shot & 3-shot \\ \hline
\multirow{9}{*}{Alexnet} & S+T & 50.4 & 61.2 & 50.0 & 62.4 & 50.2 & 61.8 \\
 & DANN & 57.0 & 64.4 & 54.5 & 65.2 & 55.8 & 64.8 \\
 & ADR & 50.2 & 61.2 & 50.9 & 61.4 & 50.6 & 61.3 \\
 & CDAN & 50.4 & 60.3 & 48.5 & 61.4 & 49.5 & 60.8 \\
 & ENT & 50.7 & 64.0 & 50.0 & 66.2 & 50.4 & 65.1 \\
 & MME & 57.2 & 67.3 & 55.8 & 67.8 & 56.5 & 67.6 \\
 & BiAT & 57.9 & 68.2 & 54.6 & 68.5 & 56.3 & 68.4 \\
 & APE & - & 67.6 & - & 69.0 & - & 68.3 \\
 & CDAC & \textbf{63.4} & \textbf{70.1} & \textbf{62.8} & \textbf{70.0} & \textbf{63.1} & \textbf{70.0} \\
\specialrule{.1em}{.05em}{.05em}
\end{tabular}}
\end{table}

\begin{table*}[tbp]
\centering
\caption{Accuracy(\%) of CDAC using Resnet34 as the backbone on \textit{DomainNet} under the setting of 3-shot. In the UDA setting, the supervised cross-entropy loss $\bm{L}_{\bm{CE}}$ refers to the model trained only with labeled source samples.}
\label{ablation}
\resizebox{\linewidth}{!}{
\begin{tabular}{c|c|l|cccc|ccccccc|c}
\specialrule{.1em}{.05em}{.05em} 
Net & \multicolumn{2}{c|}{Setting} & $\bm{L}_{\bm{CE}}$ & $\bm{L}_{\bm{AAC}}$ & $\bm{L}_{\bm{PL}}$ & $\bm{L}_{\bm{Con}}$ & R$\rightarrow$C & R$\rightarrow$P & P$\rightarrow$C & C$\rightarrow$S & S$\rightarrow$P & R$\rightarrow$S & P$\rightarrow$R & Mean \\ \hline
\multirow{11}{*}{Resnet34} & \multicolumn{2}{c|}{UDA} & \Checkmark & & & & 57.8&	61.4&	58.1&	53.4&	58.2&	49.6&	72.3& 58.6 \\ 
 & \multicolumn{2}{c|}{UDA} & \Checkmark & \Checkmark & & & 64.6 & 64.8 & 64.8 & 59.9 & 62.2 & 58.8 & 73.0 & 64.0 \\
 & \multicolumn{2}{c|}{UDA} & \Checkmark & & \Checkmark & & 68.0 & 73.2 &	68.3 & 61.8 & 67.0 & 63.1 &	76.5 & 68.2 \\
 & \multicolumn{2}{c|}{UDA} & \Checkmark & \Checkmark & \Checkmark & & 76.9 & 73.9 & 73.9 & 66.2 & 70.2 & 69.0 & 79.3 & 72.8 \\
 & \multicolumn{2}{c|}{UDA} & \Checkmark & \Checkmark & \Checkmark & \Checkmark & 77.1 & 74.4 & 73.2 & 67.0 & 70.4 & 69.6 & 79.6 & 73.0 \\ \cline{2-15} 
 & \multicolumn{2}{c|}{SSDA} & \Checkmark & & & & 60.0 & 62.2 & 59.4 & 55.0 & 59.5 & 50.1 & 73.9 & 60.0 \\
 & \multicolumn{2}{c|}{SSDA} & \Checkmark & \Checkmark & & & 69.4 & 68.1 & 68.3 & 62.8 & 65.6 & 62.0 & 76.9 & 67.6 \\
 & \multicolumn{2}{c|}{SSDA} & \Checkmark & & \Checkmark & & 76.7 &	73.6 &	76.3 &	66.9 &	70.3 &	69.3 &	80.4 &	73.4 \\
 & \multicolumn{2}{c|}{SSDA} & \Checkmark & \Checkmark & \Checkmark & & 78.7 & 74.9 & 78.5 & 69.7 & 73.2 & 71.1 & 81.6 & 75.3 \\
 & \multicolumn{2}{c|}{SSDA} & \Checkmark & \Checkmark & \Checkmark & \Checkmark & 79.6 & 75.1 & 79.3 & 69.9 & 73.4 & 72.5 & 81.9 & 76.0 \\
\specialrule{.1em}{.05em}{.05em} 
\end{tabular}
}
\vspace{-2.0mm}
\end{table*}

\section{Experiments}

\subsection{Setups}

{\bf Benchmark datasets:} We evaluate the efficacy of our proposed CDAC approach on several standard SSDA image classification benchmarks, including the \textit{DomainNet}\footnote {http://ai.bu.edu/M3SDA/}~\cite{peng2019moment}, \textit{Office-Home}\footnote{http://hemanthdv.org/OfficeHome-Dataset/}~\cite{venkateswara2017deep} and \textit{Office}\footnote{https://people.eecs.berkeley.edu/~jhoffman/domainadapt/}~\cite{saenko2010adapting}. 
\textit{DomainNet} is initially a multi-source domain adaptation benchmark, and MME~\cite{saito2019semi} borrows its subset as one of the benchmarks for SSDA evaluation. Similar to the setting of MME, we only select 4 domains, which are Real, Clipart, Painting, and Sketch~(abbr. {\bf R}, {\bf C}, {\bf P} and {\bf S}), each of which contains images of 126 categories. \textit{Office-Home} is a widely used UDA benchmark and consists of Real, Clipart, Art and Product~(abbr. {\bf R}, {\bf C}, {\bf A} and {\bf P}) domains with 65 classes. \textit{Office} is a relatively small dataset contains three domains including DSLR, Webcam and Amazon~(abbr. {\bf D}, {\bf W} and {\bf A}) with 31 classes. For fair comparisons, the settings of our benchmark datasets refer to the existing SSDA approaches~\cite{saito2019semi, qin2020opposite, kim2020attract}, including adaptation scenarios of each dataset, the number of labeled target data~(typically 1-shot or 3-shot per class), sample selection strategies, etc.

{\bf Implementation details:} Similar to previous SSDA work~\cite{saito2019semi, 2020Bidirectional}, we choose Alexnet and Resnet34 as our backbone networks. Firstly, the feature extractor is initialized with a pre-trained model on ImageNet\footnote{http://www.image-net.org/} and the linear classification layer is initialized randomly, which has the same setting as~\cite{ saito2019semi, qin2020opposite, kim2020attract, 2020Bidirectional}, such as architecture, output feature size, and so on. To balance multiple loss terms, we set $\lambda$ in Equation (\ref{final_loss}) to $1.0$ and $\nu$ in Equation (\ref{L_con}) to $30.0$. Also, we set the confidence threshold $\tau=0.95$ in Equation (\ref{L_pl}). We implement our experiments on the widely-used PyTorch\footnote{https://pytorch.org/} platform. Additionally, in each iteration, we first train our model with the standard cross-entropy loss only on labeled data from both source and target domains and then add our proposed losses on unlabeled target data to further optimize the model. Furthermore, we introduce RandAugment~\cite{2019RandAugment} as the data augmentation techniques used in this work. Finally, for fair comparisons, other experimental settings in our proposed CDAC, such as the optimizer, learning rate, mini-batch size, are the same as MME~\cite{saito2019semi}. Our code is publicly available at https://github.com/lijichang/CVPR2021-SSDA.

{\bf Baselines:} We compare CDAC with previous state-of-the-art SSDA approaches, including ``{\bf MME}'' \cite{saito2019semi}, ``{\bf UODA}'' \cite{qin2020opposite}, ``{\bf BiAT}'' \cite{2020Bidirectional}, ``{\bf Meta-MME}'' \cite{li2020online}, ``{\bf APE}'' \cite{kim2020attract}, ``{\bf S+T}'', ``{\bf DANN}'' \cite{2017Domain} and ``{\bf Ent}'' \cite{Grandvalet2005Semi}. Specifically, the model of the ``S+T'' method is trained using labeled source and target data only. In addition, ``DANN'' and ``Ent'' are both representative UDA methods and we re-train the models of ``DANN'' and ``Ent'' with an additional supervision loss by adding a few labeled target data.


\subsection{Comparisons with the state-of-the-arts}

Results on \textit{DomainNet}, \textit{Office-Home} and \textit{Office} under the settings of 1-shot and 3-shot with Alexnet and Resnet34 as backbone networks are reported in Table~\ref{base_domainNet_table},~\ref{base_office_home_table} and~\ref{base_office_table}, respectively. As illustrated, our proposed CDAC significantly outperforms the state of the art throughout all experiments. 

{\bf On DomainNet:} As shown in Table~\ref{base_domainNet_table}, our CDAC significantly outperforms the existing approaches in all adaptation scenarios on \emph{DomainNet}. Using Alexnet as the backbone, our method surpasses the existing best performing approach by $6.6\%$ and $6.8\%$ on average w.r.t the 1-shot and 3-shot settings respectively. Compared with the competing approaches using Resnet34 as the backbone, CDAC also achieves the best results in all cases and surpasses the current best results by $6\%$ and $4.3\%$ in the settings of 1-shot and 3-shot. Note that ``MiCo'' proposed in~\cite{yang2020mico} is an unpublished work concurrent with ours, and the average performance of our CDAC using ResNet34 as the backbone is $0.4\%$ higher than ``MiCo'' under the 3-shot setting. 

{\bf On Office-Home and Office:} To be consistent with the previous methods and achieve a fair comparison, we just employ Alexnet as the backbone on the \textit{Office} benchmark. As shown in Table~\ref{base_office_home_table} and Table~\ref{base_office_table}, our CDAC outperforms all comparison methods w.r.t mean accuracy on both datasets. In addition, it is worth noting that our method using Alexnet as the backbone achieves superior performance for most adaptation scenarios on \textit{Office-Home}, and consistently achieves the best performance on \textit{Office} w.r.t the adaptation scenarios of both ``W$\rightarrow$A'' and ``D$\rightarrow$A''.

\subsection{Analysis}
\begin{figure}[t]
\begin{center}
\includegraphics[width=8.0cm,height=3.5cm]{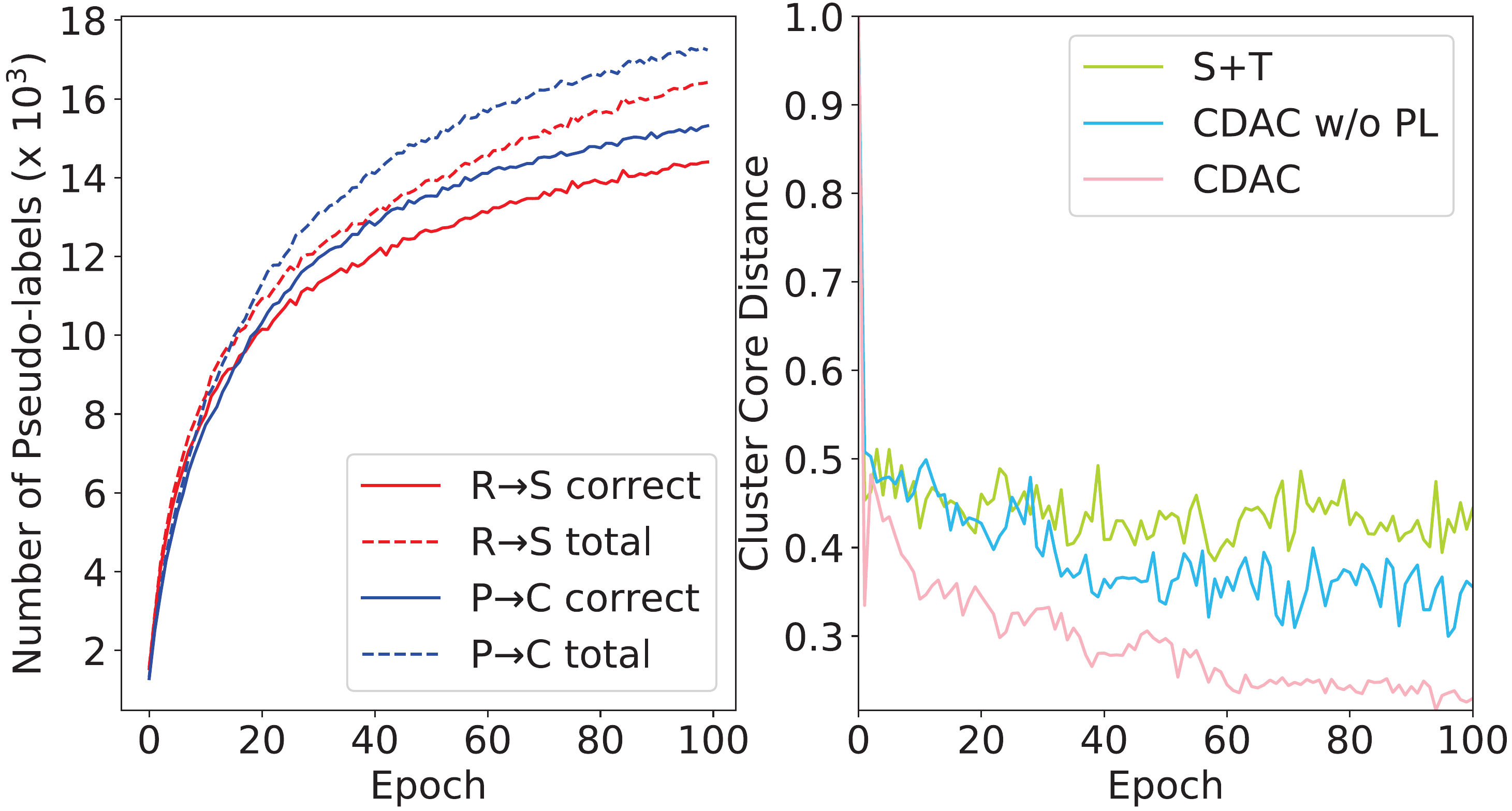} 
\end{center}
\caption{{\bf Left: }The quantity and correctness of the proposed pseudo labeling technique on two adaptation scenarios of~\textit{DomainNet} (i.e., ``R$\rightarrow$S'' and ``P$\rightarrow$C''), using Resnet34 as the backbone under the setting of 3-shot and 1-shot, respectively. {\bf Right: }Variation of Cluster Core Distance among different approaches during model training while class ``axe" is taken as an example.}
\label{fig:ccd_variation_and_pseudolabeling}
\end{figure}

 {\bf Ablation studies:} We perform ablation studies on both SSDA and UDA settings to analyze the effectiveness of each loss term in our proposed CDAC, including $\bm{L}_{\bm{CE}}$, $\bm{L}_{\bm{AAC}}$, $\bm{L}_{\bm{PL}}$ and $\bm{L}_{\bm{Con}}$. All experiments are conducted on \textit{DomainNet} using Resnet34 as the backbone under the 3-shot setting. As shown in Table~\ref{ablation}, we regard the model trained with the cross-entropy loss $\bm{L}_{\bm{CE}}$ only on labeled samples from both domains as the baseline for SSDA. And then, by combining both $\bm{L}_{\bm{AAC}}$ and $\bm{L}_{\bm{PL}}$ with $\bm{L}_{\bm{CE}}$, the trained model achieves $25.3\%$ higher average performance than the baseline, while the classification accuracy is on average $17.6\%$ ($+\bm{L}_{\bm{AAC}}$) or $23.4\%$ ($+\bm{L}_{\bm{PL}}$) higher than the baseline when only one of them is used together with the cross-entropy loss. Furthermore, the model trained with all loss functions reaches the best classification performance compared with the baseline. Moreover, each loss term proposed in our approach used for unlabeled target examples also shows similar roles in improving classification performance under the UDA setting.
 
{\bf Effectiveness of Adversarial Adaptive Clustering:} 
To evaluate the effectiveness of the adversarial adaptive clustering loss, we refer to~\cite{tao2019minimax, luo2019taking} and employ Cluster Core Distance (CCD) to measure the distance between the source and target domain feature clusters within the same class. Generally speaking, the more aligned cross-domain feature clusters are, the smaller the CCDs are. We compare our CDAC model with ``S+T'' and ``CDAC w/o PL'' (a degraded version of CDAC, which is trained with only $\bm{L}_{\bm{CE}}$ and $\bm{L}_{\bm{AAC}}$). As shown in the right of Figure \ref{fig:ccd_variation_and_pseudolabeling}, it can be observed that the CCDs of all three methods decrease gradually during model training and it demonstrates that the source and target domain clusters within each class become closer. And both ``CDAC w/o PL'' and CDAC can result in better feature alignment than S+T. The CCD obtained from the model trained with CDAC finally converges to the minimum value, indicating that CDAC overall shows the best classification performance. This demonstrates the effectiveness of the proposed adversarial adaptive clustering loss in guiding the model towards learning better cluster-wise feature alignment. 

\begin{figure}[h]
\centering
\includegraphics[width=7cm,height=7cm]{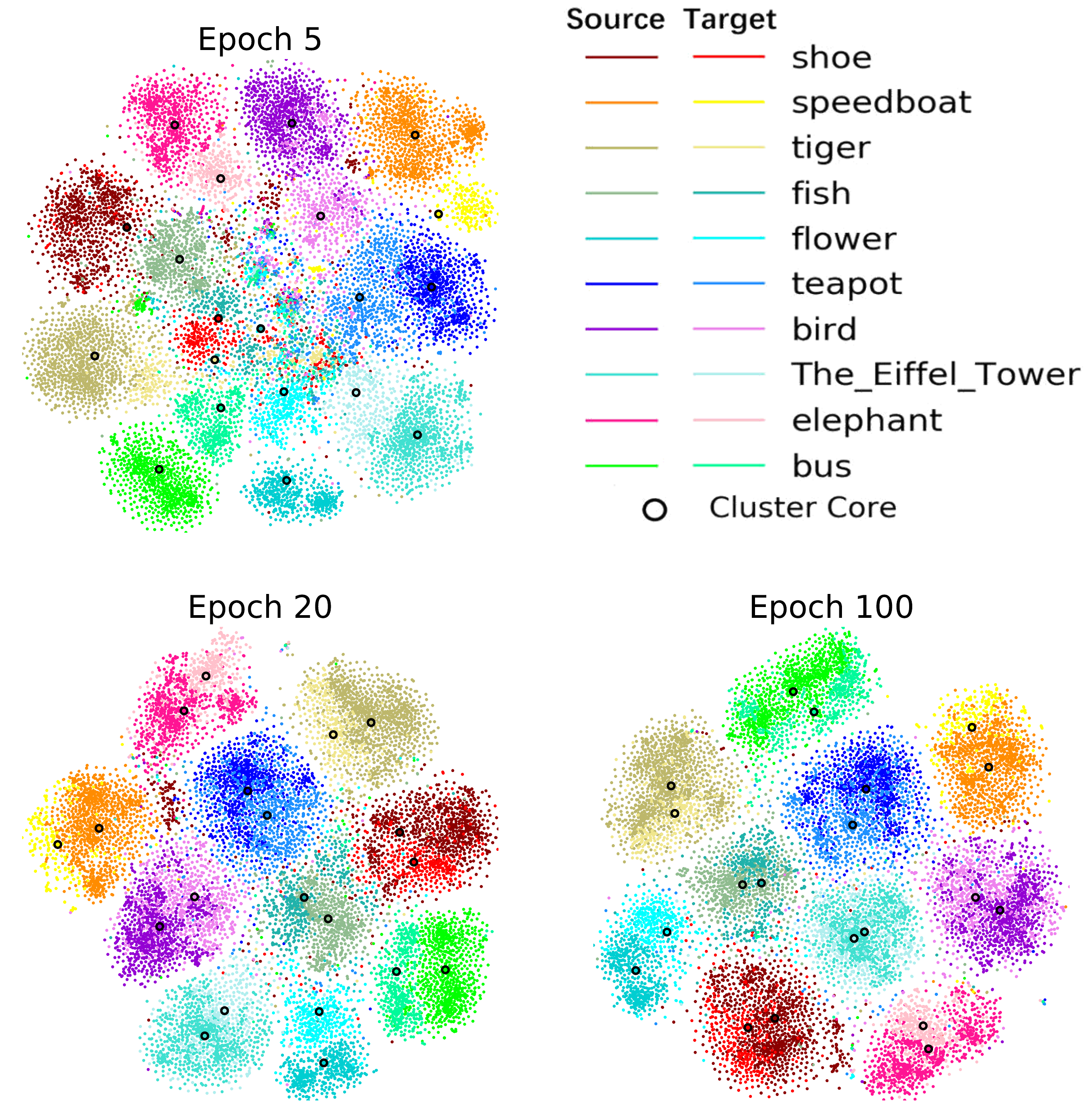}
\caption{Visualization for feature distribution variations during model training with t-SNE. We choose 10 representative classes under the adaptation scenario ``R$\rightarrow$S'' on~\textit{DomainNet} and their corresponding feature distributions from both source and target domains are displayed with different colors while black cycles represent cluster cores. We can observe that the final features have better cross-domain alignment than those at the beginning.}
\label{fig:tsne}
\end{figure}

{\bf Effectiveness of Pseudo Labeling:} 
The left subfigure in Figure~\ref{fig:ccd_variation_and_pseudolabeling} shows the quality and correctness of our proposed pseudo labeling technique in the model training process under two adaptation scenarios on~\textit{DomainNet} (i.e., ``R$\rightarrow$S'' and ``P$\rightarrow$C'' using Resnet34 as the backbone under the setting of 3-shot and 1-shot, respectively). It displays that a large proportion of unlabeled data is given correct pseudo-labels~(up to $59.9\%$ and $63.8\%$ of total training examples per epoch at the best performance, respectively), which demonstrates the effectiveness of the proposed pseudo labeling technique in CDAC.

{\bf Feature visualization:} 
We report with t-SNE~\cite{maaten2008visualizing} to display the gradual process of cluster-wise feature alignment during model training using the adaptation scenario ``R$\rightarrow$S'' of~\textit{DomainNet} under the setting of 3-shot with Resnet34 as the backbone. As shown in Figure~\ref{fig:tsne}, we visualize the variations of the cluster and the corresponding cluster core of each class in the model training process. It can be observed that as the model optimization progresses, target features gradually converge towards target cluster cores, and each cluster in the target domain also gradually moves closer to their corresponding source cluster cores, showing a cluster-wise feature alignment effect. We take the ``bus" class as an example. In Epoch 5, the feature distributions from both source and target domains are relatively far away. Then, as the model iterates, they gradually approach and finally achieve a perfect match at the last epoch. 

\section{Conclusions}
We have presented a novel approach called Cross-domain Adaptive Clustering~(CDAC) to solve the SSDA problem. CDAC consists of an adversarial adaptive clustering loss to guide the model training towards grouping the features of unlabeled target data into clusters and further performing cluster-wise feature alignment across domains. Furthermore, an adapted version of pseudo labeling is integrated into CDAC to enhance the robustness and power of cluster cores in the target domain to facilitate adversarial learning. Extensive experimental results, as well as ablation studies, have validated the virtue of our proposed method.

\section*{Acknowledgements}
 This work was supported in part by National Key Research and Development Program of China~(No.2020YFC2003900), in part by the Guangdong Basic and Applied Basic Research Foundation~(No.2020B1515020048) and in part by National Natural Science Foundation of China~(No.61976250 and No.U1811463). This work was also supported by Meituan.

{\small
\bibliographystyle{ieee_fullname}
\bibliography{egbib}
}

\end{document}